# Rolling in the deep of cognitive and AI biases

NICOLETA TANTALAKI, Aristotle University, Greece
ATHENA VAKALI, Aristotle University, Greece

Artificial Intelligence (AI) currently acts solo or as a human companion in decisions made at several sensitive domains, like healthcare, finance, and law. AI systems, even those carefully designed to be fair, are heavily criticized for delivering misjudged and discriminatory outcomes against individuals and groups of people. The continuous unfair and unjust AI outcomes, indicate that the significant impact of human and societal factors on AI biases, is currently overlooked. It is now urgent to view and understand AI as a sociotechnical system, inseparable from the conditions in which it is designed, developed and deployed. This work addresses this critical issue by proposing a systematic methodology under which human cognitive biases intertwine within the overall AI lifecycle. By identifying how harmful human actions influence AI's biases, we reveal human-to-AI biases hidden pathways, leveraging on the major human heuristics, as identified in cognitive science. Central to this effort is a mapping approach that systematically connects human heuristics to specific AI biases, uncovering meaningful patterns of human–AI influence and interdependence that shape fairness outcomes. We envision that this work will inspire a genuine human-centric AI fairness approach by revealing hidden biases cause and effects.

## KEY INSIGHTS

- AI is never developed in isolation; it is deeply shaped by the ways humans think, decide, and interact with the world. Cognitive shortcuts and implicit heuristics embedded in human decision-making often migrate into AI design, influencing system behavior and leading to unfair or unintended outcomes.
- `HAI-ROLL` framework systematically identifies AI lifecycle actions that act as sources of AI bias, links them to the underlying human heuristics, and reveals hidden relationships between human (cognitive) processes and algorithmic biases.
- By surfacing these pathways, `HAI-ROLL` opens new insightful and practical AI design guidance. It promotes bias-aware decision-making across the AI lifecycle and reframes fairness not as a technical afterthought but as a human-centered responsibility woven into the entire AI development process.

## 1 INTRODUCTION

Artificial intelligence (AI) has rapidly transformed decision making in numerous domains. Acting alone or as a human companion, AI offers health suggestions, approves loans, detects criminal profiles, distributes digital content, and supports a wide range of additional decision-making tasks. [30]. However, the hypothesis that AI would result in more objective and rational decisions than humans is not verified. In numerous real-world cases, AI algorithms have exhibited significant issues, caused by unexpected and unpredictable biases hidden within opaque systems [27]. Since AI reflects not only the inequitable patterns present in society, but also the fair or biased perspectives of its creators [14], it is evident that a sociotechnical bias exploration is urgently needed [30].

In pursuit of *Fairness in AI (FairAI)*, computational methods like re-sampling or re-weighting are often inadequate, costly, and rely on unrealistic assumptions—such as prior knowledge of which sensitive features are most bias-prone [18]. A rich and emerging bibliography has detected specific types of bias in all phases of the AI lifecycle of data collection (*pre-processing*), modeling (*in-processing*) and deployment (*post-processing*), along with a a divergent set of computational fairness metrics for bias mitigation [14, 26, 35, 37]. Although such studies have shown some benefits, AI (computational) biases remain fierce and harm minority or underprivileged

Authors' Contact Information: Nicoleta Tantalaki, nicoleta@uom.gr, Athena Vakali, avakali@csd.auth.gr

populations. The rapid and widespread adoption of Large Language Models (LLMs) has intensified the need to identify hidden and underexplored forms of bias. This is particularly relevant in models fine-tuned through methods like Reinforcement Learning from Human Feedback (RLHF), which aim to produce outputs that are truthful, helpful, and harmless[8]. However, because human feedback reflects subjective values and social norms, this alignment process can unintentionally reinforce cultural biases, dominant perspectives, or inconsistent expectations [8, 20].

Our work is highly motivated by the fierce AI biases and the need to understand the causes and effects of the biases, largely perplexed in our new symbiotic reality of humans - AI. Humans make decisions driven by their own perception of reality, and such *human heuristics* (mental shortcuts, motivations, time constraints or other limitations) often result in systematic errors, known as *cognitive* biases [36]. It is impressive that long-time seminal work in cognitive science has identified more than 180 cognitive biases in human judgment and this list is continuously evolving [2]. Unfortunately, such valuable cognitive science evidence is up to now overlooked in AI, and we lack a *deep* exploration among the human heuristics (cognitive) and the AI (computational) biases interdependencies [37]. We claim that we must put forward the human-side factors, which cause or steer the many AI biases [25, 30], under a new line of sociotechnical research to prevent harms to minority or underprivileged populations (gender, age, ethnicity, vulnerability, etc.) [31, 37].

Unlike prior frameworks that focus solely on computational bias, our work traces the roots, forms, and manifestations of cognitive and AI biases. We introduce a systematic mapping framework that reveals interdependencies between human heuristics and AI biases across the AI lifecycle. This cognitive-to-AI bias mapping establishes a human-centric approach to FairAI that: (a)*identifies particular actions as sources of AI (computational) biases*; (b)*rolls in the deep of these actions causes to uncover their inherent cognitive (human-side) heuristics, which are projected at each of the AI lifecycle phases*; (c)*introduces a systematic methodology to reveal cognitive heuristics and AI biases inter-relationships, based on the evidence of a flexible exploratory analysis*. Our work relies on the scientific ground of human heuristics introduced by the Nobelist Daniel Kahneman and pioneer researchers [16, 22, 23, 36], and on the most state of the art prominent AI biases [14, 24, 26, 35, 37]. By bridging insights from cognitive science and AI research, we open a new dialogue for systematic, cross-disciplinary exploration of bias in AI systems. Ultimately, we aim to equip practitioners with a clear understanding of which decision points are most susceptible to cognitive influence, offering structured support for reflection, anticipation, and bias-aware planning and deployment. Our work provides a theoretical foundation for human-centric, bias-aware decision making through a cognitive reflection framework designed to guide both practitioners and policymakers, while supporting future empirical studies.

## 2 COMPUTATIONAL BIASES IN THE AI LIFECYCLE

Among the many types of computational bias explored in the AI lifecycle (pre-, in- and post-), the most prominent and widely studied [14, 24, 26, 35, 37] are: **(a) Historical bias**, when a particular selective handling of certain groups or communities is advantaged or favored, while others historically remain disadvantaged or excluded. Such biases arise even if the data are sampled flawlessly, since selective preference may be inherent in the original data sources, as a result of the majority following existing patterns or norms [30, 36, 37]. For example, a mechanical engineering firm's dataset might have inherent historical biases if it contains a significantly higher number of male employee records. Such an imbalance stems from the firm's hiring norms favoring male candidates due to the belief that women are less capable; **(b) Representation bias**, when non-representative datasets are generated by sampling from a population, i.e. when samples are of small scale, or wrongly selected, neglecting underrepresented groups with fewer data points, resulting in the model's failure to generalize for these groups [26, 35, 37]. For example, when a recruiting company trains an AI model using a data sample primarily consisting of high-salary male employee profiles, results in biased hiring recommendations since the model cannot recognize high-salary female employees or value female candidates; **(c) Measurement bias**, when particular features and labels are chosen, collected, or computed, acting as "proxies" that poorly substitute some construct (an idea or concept) [24, 35]. For example, if an AI model uses the Body Mass Index (BMI: estimated by a rate of an individual's weight over her/his height squared) as a proxy for obesity risk, it can mistakenly classify individuals with greater muscle mass as obese, since BMI cannot distinguish between muscle and fat and fails to take into account key factors (such as age, gender, or ethnicity); **(d) Algorithmic bias**, when inappropriate model design choices made mostly for a model's architecture, functions and hyper-parameters, require several mathematical abstractions. Even under non-discriminatory intent, these choices often lead to biases since they show disproportional error rates against underrepresented groups [37, 38]. For example, an underfitting model (due to its oversimplified nature) has restricted capacity in detecting important data patterns and yields low predictive accuracy, particularly for minority groups that are underrepresented in the dataset[18, 38]; **(e) Evaluation bias**, when the benchmark dataset (used for a model's evaluation), is not representative of the target population and/or inappropriate performance metrics are used [37]. For example, if an AI smile detector model is trained on a dataset lacking adequate representation of Asian female individuals and the testing benchmark is similarly unbalanced, mirroring the training set, the bias against Asian females will go unnoticed [24]; **(f) Deployment bias**, when an AI model is designed to solve a problem, and then it is used to address a different problem in practice. This often occurs when an AI model is built and evaluated to provide recommendations, but is deployed as if it was fully autonomous. In this case, human reviewers overestimate its credibility, leading to a bias known as *automation* bias [35].

Given that all of the AI phases, but also the emerging LLMs models are largely driven by human experts, we summarize humans inherent heuristics in the next section and we then introduce our methodology for mapping these human heuristics to AI biases in Section 4.

## 3 HEURISTICS AND BIASES BEHIND HUMAN DECISION MAKING

The detection of computational bias—recently characterized as just the tip of the bias "iceberg" [30]—has been criticized for its limited ability to account for the impact of human and systemic biases throughout the AI lifecycle [14, 31, 37]. To resolve such an inefficiency, we explore the groundbreaking work about "Judgment under Uncertainty" which has revealed that the broad spectrum of cognitive biases is rooted to three core heuristics, namely *representativeness*, *availability*, and *anchoring and adjustment* [36], as well as the "Heuristics and Biases: The Psychology of Intuitive Judgement" work which has highlighted the prime role of the *affect* heuristic in human intuitive judgements [16, 22, 23]. These heuristics are briefly summarized below, serving as a foundation for the following section, which illustrates how these human cognitive tendencies can give rise to, or interact with, AI biases.

**Representativeness** refers to the spontaneous estimation of an event's probability, based on how closely it resembles a known situation or prototype that exists in a person's mind. People usually overweight certain information because they believe it is representative of reality, since this information is more recent, or has received more media attention, or is presented in a way that makes it seem more typical of a category. The representativeness heuristic leads to human biased views, due to stereotype beliefs [16, 22, 36]. For example, a woman who is characterized as friendly and empathetic is more likely to be labelled as a nurse than as a mechanical engineer, since her traits fit the stereotype of a person in a care-giving role.

**Availability** refers to the mental shortcuts humans take when assessing the frequency or plausibility of an event, based on how easily examples come to mind. This attachment to immediate thoughts often prohibit further exploration of potentially significant information, since more frequent instances are recalled better and faster than less frequent ones, and repeated occurrences are preferred over rare events. A prominent cognitive bias in such cases is the so called *illusory correlation* occurring when people conclude that variables are correlated because their pairings come to mind easily (quick to grasp, seem likely) [16, 22, 36]. For example, the availability heuristic occurs when people assume that a high number of women hold leadership positions based on the frequent visibility of a company's "Female Empowerment" campaign. This campaign is easily retrievable from memory due to its regular marketing exposure, leading to an overestimation of the actual engagement of women in leadership roles.

**Anchoring and adjustment** occurs when people base their estimates or predictions on an initial piece of information (which may be completely irrelevant) that becomes ingrained in their minds. This information acts as an "anchor", trapping human views to their original opinions, even if they are unrelated to a newly presented case. Such anchor dependency may lead to a critical cognitive bias known as *confirmation bias*, which refers to the human tendency to favor information that confirms initial views, making insufficient adjustments, and selectively preferring data that reinforce their initial anchor point [29, 36]. The anchoring heuristic can even influence individuals with a high level of knowledge or expertise, engaged in comparative decision-making, swayed by multiple anchors stored in their minds [22].

**Affect** refers to quick decisions based on emotions and feelings, leading to reactions that are more affectionate than logical or analytical. Kahneman's fast and intuitive so called "System 1 thinking" [22] reflects such people's tendency to jump to conclusions and skip information gathering, resulting in a cognitive bias known as *conclusion bias*. Although humans may also engage in deliberate, slow, and conscious reasoning "System 2 thinking" [22], their emotional attachments also influence them to seek information that confirms their preexisting beliefs, giving rise once again to *confirmation bias* [22, 23]. For example, the affect heuristic may manifest in hiring decisions when a recruiter who feels disappointed by previous female employees, unconsciously prioritizes male candidates as more capable leaders, despite evidence that new female candidates have equal or superior qualifications.

Given the radical and unprecedented mixed human-AI decision making, we must urgently reconsider the role of these human heuristics and respond to the next critical Research Questions: [*RQ1 : how can we effectively define a solid methodology to reveal human heuristics and AI biases inter-dependencies?*]; [*RQ2 : which human heuristics affect the most critical AI biases and when?*]

## 4 HAI-ROLL : A FRAMEWORK FOR MAPPING HUMAN HEURISTICS AND AI BIASES

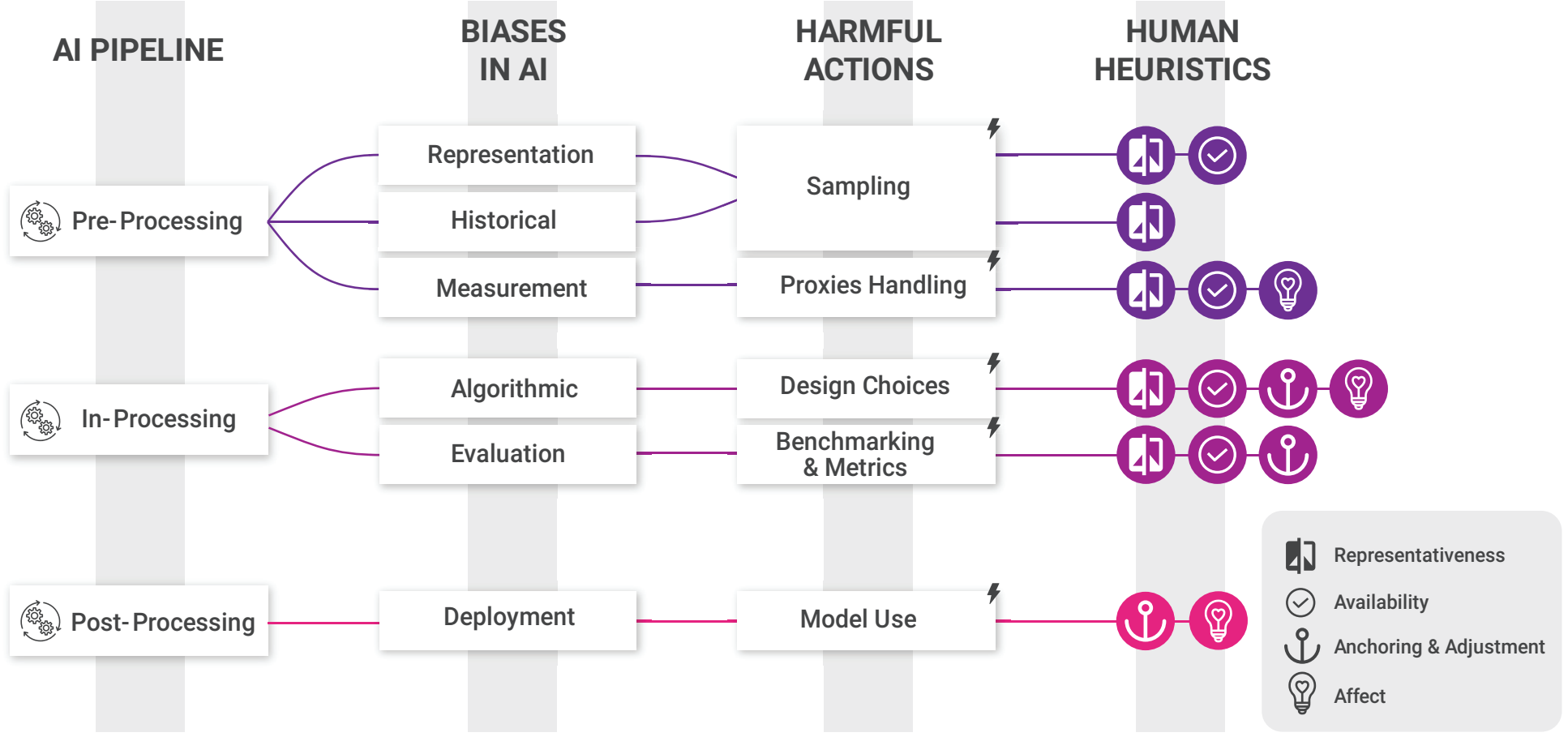

Fig. 1. Human heuristics and AI biases mapping via Harmful actions at the AI lifecycle

Recent surveys have roughly highlighted some indicative harmful actions which may lead to AI biases [14, 19, 24, 26, 35, 37]. However, the origins of these AI biases are barely and poorly referenced [19]. We deviate from this restrictive line of research by our proposed framework, so called `HAI-ROLL`, which positions AI lifecycle harmful actions as catalysts for mapping human heuristics and AI biases (Fig. 1). `HAI-ROLL` framework enables a sosciotechnical view since it : (i) identifies and encodes the particular harmful actions ([HA.xx]) that lead to unfair outcomes, at each phase of the AI lifecycle (pre-,in-, post-); (ii) detects the specific AI biases (of Section 2) that have emerged due to these [HA.xx]; and (iii) traces the origins of these [HA.xx] by exploring the well-established human heuristics (of Section 3). Our transitive reasoning responds to the above RQs since, ultimately, we *roll into the deep* evidence of human psychology to reveal the relevance, influence, and pathways linking human heuristics to AI biases. Our methodology is grounded on interdisciplinary scientific evidence not only to detect human and AI biases correlations, but also to offer a systematic and adaptive framework for innovative and human-centric fairness-by-design strategies.

## 4.1 Pre-processing Phase

Data and feature selection are critical initial actions that impact the overall AI lifecycle. Since data scales prohibit the inclusion of the entire data population, certain issues are expected to be raised. For example, some groups may be excluded or underrepresented, some features cannot be acquired, etc., leading to inaccurate and possibly harmful results (as indicated next).

**Harmful Action [HA.P1]:** ***Use of inappropriate (and non-representative) data samples*** **→ Computational bias:** ***Representation & Historical***
Several ill-choices are made when the variability of the population is not taken into consideration; the size of the sample used is not the appropriate one, the sampling method is uneven or limited [35]. Such choices often lead to *AI representation* bias, yielding results that are applicable only to a subset of the relevant population and not to the broader population itself (due to a non-representative sample). Even a "perfect" sampling plan might turn into an [HA.P1], since *AI historical* biases might pop up when datasets embed their own ingrained biases [1]. "Baked-in" biases, i.e. the ones which are due to people's prejudices, perpetuate in the AI lifecycle and discriminate mostly disadvantaged groups (e.g., indigenous populations, women, people with disabilities, the elderly, etc) [30]. A notable example of *historical* (also known as *societal*) bias is seen in a 2018 image search for "women CEOs," which returned disproportionately few female results. This wasn't due to flaws in data sampling but because only about 5% of Fortune 500 CEOs were women at the time. The algorithm simply mirrored existing societal imbalances, highlighting how bias can be embedded even in "accurate" reflections of reality [26].

- *Human Heuristics mapping*: the projection of **representativeness** heuristic at the AI **representation** bias is evident in such [HA.P1] data sampling cases. This critical issue has been largely flagged in AI face recognition algorithms which are typically trained on images of people from developed countries and consequently misidentify faces of individuals from racial groups that are underrepresented in the training data [3]. Humans consistently overestimate the proportion of the world that is similar to themselves and given that most AI agents are trained in developed countries, the selection of the sample reflects individuals who resemble the typical profile of people from those regions [3, 25]. Moreover, the **availability** heuristic often lies behind [HA.P1] sampling cases, due to human tendency in placing greater emphasis on the most readily available data, even if it does not fully represent the entire population [17, 22]. Datasets,though, may be available and seem representative but still exhibit entrenched biases. Datasets that derive from populations with baked-in **historical** biases, lead to systems that reinforce patterns of discrimination. The **representativeness** heuristic is usually behind a lot of people's bigotry as this heuristic makes people believe that a stereotype is true [30, 36]

**Harmful Action [HA.P2]: *Misuse of proxy variables* → Computational bias: *Measurement***
A proxy, i.e. a concrete measurement assigned to a variable (Sec. 2), is chosen when there is a need to approximate a "construct" (i.e a concept or an entity), which is unobservable or immeasurable. Surely, the choice of appropriate proxies for an AI model is a challenging task. Proxies have been criticized for their role in AI **measurement** biases and their inaccurate or/and discriminant classifier outcomes, since they either poorly reflect the target constructs or use different estimates between groups [37]. Especially when protected attributes (eg. gender, ethnicity) are selected as proxies, challenges emerge since such attributes partition a population into different groups that should be treated equally. A straightforward suggestion for AI developers would be to avoid utilizing protected attributes (associated with historically discriminated social groups) as features in the training set, towards deploying AI models which do not rely on sensitive information and thus do not yield biased outcomes (**direct discrimination**). Unfortunately, scientific evidence has shown that even removing such protected-sensitive features does not prevent discrimination, as latent variables can be inferred implicitly (e.g., gender has been reported to be inferred through browsing history, and ethnicity can be inferred by a neighborhood zip code) [30, 37]. Such **indirect discrimination** has occurred in scenarios [HA.P2] that set neutral or objective goals but end up treating groups or individuals less favorably, based on practices that disadvantage a protected group.

- *Human Heuristics mapping:* Feature selection is made mostly by humans(e.g AI developers) who might choose the features which mostly align with their own beliefs. Thus, each AI developer's own **representativeness** heuristic may influence their assumptions about the suitability of a proxy to reflect the actual (non measurable) variable [10]. Similarly, the **availability** heuristic may cause preference for proxies that seem related to the target variable, though this perceived relationship doesn't always imply a true close correlation. Hidden variables may influence both the proxy and target variable, distorting their connection due to **illusory correlation** (identified in Sec. 2 as an AI availability bias) [36]. For example, a hiring manager might incorrectly assume that years of experience is an ideal proxy for predicting job performance. However, such a choice can reinforce gender bias. Men have historically had greater access to career paths and senior roles, which allowed them to accumulate more experience than women. Additionally, **confirmation** bias can also lead to the application of different measurement standards across groups, resulting in skewed outcomes. For instance, a fraud analyst may subject certain groups to more rigorous testing, which produces more positive cases and reinforces their initial beliefs, feelings, and prejudgements, ultimately distorting the true frequency of fraud in the overall population[38].

## 4.2 In-processing

At this phase, the AI model is trained to optimize a specified objective (e.g. maximize a reward function in RLHF or minimize a loss function). Up to now, AI model's efficiency has been challenged by using available (benchmark) datasets (different from the training data), and measuring mostly its accuracy. However, the emerging deployment of AI in sensitive and critical domains has put forward a range of important desiderata, so apart from accuracy, AI models must also be challenged for their capacity to preserve fairness and privacy [18, 42]. Thus, beyond the existing benchmarks used to estimate model's generalizability, it is now urgent to validate AI models under metrics that will assess their capacity to comply with legal and ethical rules [28].

**Harmful action [HA.I1]: *Make inappropriate model design choices* → Computational bias: *Algorithmic***
AI developers are responsible for critical model design choices, ranging from selecting the model's architecture and objective function to configuring hyperparameters (i.e., external parameters such as the number of neural network layers or the training duration). [HA.I1] cases give rise to different algorithmic bias, primarily due to the following design-related pitfalls:
*(i) Choice of the model architecture and objective function*: Simple, underfitting models may fail to capture unique

data patterns, while overfitting models, although closely fitting the training data, may adapt to noisy or unrepresentative training data [18, 38]. Model design choices become even more complex in the context of LLMs, due to their intricate architectures and multi-phase training pipelines, which can embed subtle forms of bias and amplify cognitive distortions [20]. In particular, fine-tuning techniques, originally developed to align these models (LLMs) with human values, have garnered substantial attention as methods for adapting generative systems by modifying their objective function. A key approach in this process is reward modeling (RM), in which a separate reward model is trained to approximate human preferences [8]. During reinforcement learning, this reward model effectively shapes the objective function that the main model is trained to optimize. However, because human judgments are influenced by cognitive heuristics and contextual biases, the reward model may end up favoring responses that appear more familiar, confident, or emotionally appealing, rather than those that are most accurate or helpful. As a result, the final model is optimized to reproduce these imperfect human preferences [8, 9, 20].
*(ii) Balancing multiple model requirements*: Achieving high *accuracy* while also ensuring *fairness* and *privacy* involves inherent trade-offs [18, 42]. For example, in cases where a loss function—typically minimized to reduce the difference between the algorithm's output and the ground truth—is biased toward a specific demographic group, the resulting model may perform better for that group in terms of *accuracy*. However, this improvement does not extend to other critical parameters, such as *fairness* [18, 42]. Furthermore, in privacy-preserving analysis, a cutting-edge solution known as differential privacy enhances *privacy* by using techniques such as noise injection that adds controlled noise to the data or output, masking the contribution of any individual in the dataset. However, this approach can disproportionately reduce the model's *accuracy* for underrepresented subgroups, thereby affecting the model's *fairness* [42].
*(iii) Selection of more subtle parameters*: Fine-grained design choices, such as the duration of the training process or the learning rate, can significantly affect error rates, especially for underrepresented features, which may be disproportionately impacted by such adjustments. For example, early stopping (a technique used to prevent overfitting while preserving model accuracy) may inadvertently exacerbate biases. Since underrepresented features are typically learned later in the training process, selecting an inappropriate early stopping point can negatively impact specific subsets of the data population [18].

- *Human Heuristics mapping:* AI developers make subtle choices, seek patterns, and often substitute complex problems with simpler ones guided by **representativeness** and **availability** heuristics [21]. For instance, the human preference for simpler models with fewer, more accessible parameters is linked to the **availability** heuristic [29, 30, 36]. Moreover, the **affect** heuristic can contribute to [HA.I1] cases and act as a source of **algorithmic** bias. For example, a developer may prioritize *privacy* in a healthcare system(based on prior experience or personal values) emotionally emphasizing data protection while neglecting *accuracy* concerns [18]. In the case of LLMs, empirical work showed that cognitive heuristics are already instilled in them through their training data and in-context learning patterns [32]. For example, frontier LLMs exhibit the **representativeness** heuristic by mimicking frequent training patterns, even when these conflict with sound reasoning. Their strong in-context learning capabilities can also amplify the **anchoring** effect, whereby early prompt tokens disproportionately influence the model's output [32]. RLHF-aligned models have been shown to exhibit even greater bias than their base counterparts [20]. Human annotators who tend to up-score responses that align with their own beliefs or feel more agreeable may introduce **affective** and **confirmation** bias into the reward model, thereby fostering sycophantic behavior in the aligned model (i.e., producing overly agreeable responses even when they are less factual)[33]. Similarly, human raters tend to favor longer outputs (length anchoring), prompting the model to equate verbosity with quality [34]. Ultimately, human heuristics cause models to favor agreeable or verbose outputs over factual or fair ones. Once embedded in the reward model, these biases are amplified during fine-tuning as the model learns to optimize for flawed human preferences [8, 20, 33, 34]. Furthermore, as models grow in size and capability, they become more susceptible to internalizing human-like cognitive biases, due to their increased capacity to capture and reproduce subtle patterns (including flawed reasoning) from training data and human feedback. This enhanced ability to mimic both rational and irrational human behavior creates a trade-off: greater fluency and alignment with human preferences, but also a deeper entrenchment of cognitive distortions. [20].

**Harmful Action [HA.I2]:** ***Use of inappropriate (and non-representative) benchmarks and performance metrics*** **→ Computational bias:** ***Evaluation***

In this AI phase, performance is assessed by using unseen data, and biased results might occur when such testing datasets are non-representative of the actual population. To prevent [HA.I2] cases, the overall population must be represented and the AI model should not perform well only for a subset of the population. For example, a Named Entity Recognition (NER) model trained on data with more male names, will perform well on a male-dominant benchmark dataset, but once a more balanced gender representative benchmark dataset is used, female mis-classification will be revealed. Moreover, AI models' evaluation based on metrics like accuracy or precision is heavily criticized, since these metrics don't address fairness across groups [28]. Subgroup validity approaches aim to ensure fairness by examining performance metrics across different subgroups. They use protected attributes to provide analysts with essential group information, indicating whether an individual belongs to a protected or unprotected group [28, 35]. However, generalizing a fairness definition and mathematically representing it remains too complex task since the numerous fairness metrics proposed up to now cannot be satisfied simultaneously and do not explore human-centric sources which might result in emerging AI biases [28].

- *Human Heuristics mapping:* Since humans (AI developers) make the decision on how to split the overall dataset into training and testing benchmark sets, their choices might cause [HA.I2] faults, due to their **representativeness** and **availability** heuristics (as in the [HA.P1] above). Prominent studies have also reported that the human-side **confirmation bias** in AI software testers leads them to prioritize evaluation schemes with testing datasets and metrics that confirm, rather than question the correct function of an AI model [5, 29]. Guided by their **anchoring and adjustment** heuristic, AI human testers often get stuck in their belief that the algorithm is accurate and focus on scenarios or inputs that are consistent with their initial assumptions, rather than exploring a broader range of possibilities that could challenge their initial belief [5, 25].

### 4.3 Post-processing

In this AI phase, model outcomes are utilized and interpreted, and the model is refined on the basis of user reactions. This iterative nature of AI triggers "feedback loops," where a model's output serves as a new input for further refinements (e.g., by retraining) and re-evaluation of the AI lifecycle. These loops aim to make AI models' adaptive and dynamic under varying conditions [25, 37]. However, reliance on previous AI outputs poses the risk of perpetuating biases, leading to what's known as the "butterfly effect" [15], where the origins of bias become unclear and untraceable and complicate AI fairness resilience.

**Harmful Action [HA.D1]:** ***Use of the AI model at scenarios which deviate from its initial scope*** **→ Computational bias:** ***Deployment***

In the AI validation phase, the original intended scope of the AI model must be carefully considered and its actual use must comply with it [30]. The *deployment* bias spotted in this phase occurs due to [HA.D1] cases which highlight discrepancies in the AI's model use, when compared to the original scope for which the model was designed. For example, the wide range of AI models, from strictly automated decision making without human involvement to AI that supports human decision makers [19], embeds [HA.D1] risky cases. Such risks emerge when humans "blindly" put their faith in AI automated decision making, whilst the AI model was built only for a human-supportive scope [19]. This phenomenon of anchoring bias in AI-assisted setting is known as *automation* bias [29].

- *Human Heuristics mapping:* The **anchoring and adjustment** heuristic lies behind several [HA.D1] cases which cause **deployment** bias [29, 30]. **Confirmation** bias occurs when decision makers favor AI-generated results that support their pre-existing beliefs (anchors), forcing them to disregard other potentially more relevant or accurate results that may contradict their assumptions [29]. Decision makers often fall in [HA.D1] faults, by relying on AI decision-support systems, even without fully understanding their inner-works. Several studies have highlighted that domain experts who make decisions tend to overestimate their understanding of a domain but they are not aware of the underlying (opaque) AI and data science tasks which impact their decisions [22, 30]. The **affect** heuristic has also been identified as a driver of [HA.D1] cases, since the more human-like a system appears, the more likely it is that users attribute more human traits and capabilities to it. Automated systems often foster a sense of security, easing anxiety and increasing trust, and such emotional comfort overshadows critical evaluation [29, 30, 39]. A common example of such cases is the so-called AI anthropomorphic conversational agents, i.e., LLM applications that engage users through dialogue. When people interact with LLMs, they may start to perceive them as human-like, share more information than they normally would, and rely on them more than they should [39].

## 5 HAI-ROLL IN PRACTICE : IMPLEMENTATION GUIDELINES AND LIMITATIONS

Adopting the HAI-ROLL framework in real-world AI systems demands cautious configurations and deep understanding of its benefits and limitations. Given that it is the first framework (according to authors knowledge) which rolls in the deep from human to AI bias exploration, HAI-ROLL balances the societal (human) and the technical (AI) constraints and differences. AI professionals (system designers and developers) are more aware of biases embedded in data or algorithms, and proceed with their choices based on existing fairness metrics and mitigation techniques [2, 6, 14, 28, 37]. In contrast, they are less conscious of their own cognitive biases impact in shaping their decisions during the AI pre-, in- and post- processing phases [6]. HAI-ROLL offers AI developers an actionable guideline service to critically reflect on how cognitive (often unconsciously) heuristics shape AI lifecycle technical choices. At the same time, HAI-ROLL offers actionable guidance to policymakers who will recognize the importance of a complete sociotechnical AI framework. HAI-ROLL supports AI developers to anticipate where biases may emerge and take preemptive action, while they reinforce their bias-aware critical thinking encouraging them to shift to a reflective engineering mindset (grounded in both technical evidence and psychological insights). In summary, HAI-ROLL offers a paradigm shift and revisits the human and AI synergy at each of the real world AI lifecycle phases, with the next indicative forward steps.

**HAI-ROLL as a Pre-processing driver,** since developers may identify cognitive distortions and blind spots (overlooked gaps where certain subpopulations, contextual nuances, or relevant attributes are misrepresented or excluded), which lead to representational and historical biases at data collection choices [4]. By encouraging cognitively driven data set audits (e.g., through checklists [7]) and structured documentation (e.g with datasheets [14]), HAI-ROLL supports more inclusive, context-sensitive, and cognitive-aware design and enhances bias anticipation [2, 4, 14]. It further supports proper proxy variables selection (during feature engineering), by overcoming the current eXplainable AI (XAI) methods (e.g. SHAP and LIME), which indicate feature importance, but fail to explain why particular proxies were chosen or assess their social and ethical validity [12]. HAI-ROLL highlights that explanations provided must align with the reasoning strategies and humans heuristics [29], and reveals how cognitive biases (such as illusory correlation and confirmation) can distort feature selection, prompting developers to critically examine the rationale behind their design decisions.

**HAI-ROLL as an In-processing monitor,** by safeguarding socially fair or ethically sound AI methods which currently focus on only technical optimizations (e.g. hyper-parameter tuning, adversarial training) mostly used to boost accuracy and robustness [14]. HAI-ROLL encourages AI professional teams to reflect on whether their AI Model design defaults are driven by fairness considerations or merely by AI in-processing simplicity. The emerging alignment techniques (like RLHF) in LLMs and the new forms of bias underscore the role of HAI-ROLL to support bias mitigation strategies [20]. By addressing human and AI biases concerns, HAI-ROLL offers insights to rethink ways of collecting human feedback, which must include diversifying annotator perspectives [9]. Given that RLHF reward models may easily reinforce responses that are emotionally appealing or agreeable, rather than logically sound [8, 33], HAI-ROLL introduces a complementary and practical layer for new sociotechnical insights to better train annotators and guide reward model audits. By explicitly identifying the heuristics that often shape human preferences (such as affect and confirmation), HAI-ROLL strongly supports developers to anticipate where feedback may introduce cognitive distortions into the reward model. During the AI model evaluation phase, cognitive biases can distort both the selection and interpretation of benchmarks and metrics [11]. Empirical audits have revealed frequent inconsistencies in standards and overly narrow definitions of fairness [4, 11]. Evaluation methods in real-world scenarios are often constrained by questionable assumptions (such as the availability of reliable sensitive attributes or clear group boundaries) which can lead to evaluation bias and undermine fairness guarantees [4].HAI-ROLL acknowledges how human reasoning and contextual AI settings impact fairness judgments. It offers a novel socio-technical approach to in-processing monitoring by enabling bias-aware tools (e.g., mind maps and checklists), which will facilitate more reflective evaluation practices and inspire new AI professionals and policy making synergies.

**HAI-ROLL as a Post processing auditor,** by addressing AI fairness mitigation as a shared responsibility, and as collective and collaborative synergy between AI developers, ethicists, domain experts, and, crucially, policymakers. In practice, deployment bias occurs when AI systems (despite careful design) are misused and are over-relied on, or are misunderstood by real-world adopters, since cognitive biases persist, even when explainability tools exist [4, 6, 29]. For example, displaying confidence scores may lead to over-trust and blind acceptance of flawed outputs [29], since AI accuracy and transparency alone don't ensure fairness, and don't anticipate human heuristics. In real-world AI implementations, humans rely on cognitive shortcuts and they may rubber-stamp low-quality outputs or gradually stop questioning the system as their trust increases. Such unexamined decisions may feed back into AI training data (pre-processing phases), and thus perpetuate potential discrimination through AI [29, 37]. HAI-ROLL can assist in reflecting such biases to specific metrics, like a 'blind accept' rate, that will enable inter-disciplinary teams (AI developers along with policy makers) to build dashboards, trigger justification prompts and set alert thresholds to detect drift early. HAI-ROLL may also be adopted into certification processes, to set the requirements that will promote fair and accountable AI development. For example, under Article 9 of the EU AI Act, providers must implement continuous risk management [13]. Based on the HAI-ROLL framework, a Cognitive Bias Impact Assessment (CBIA)—similar to Data Protection Impact Assessments (DPIAs) under the GDPR [41]—could be appended to this process, requiring teams to identify which cognitive heuristics are likely to surface at each stage and to document appropriate countermeasures. In addition, HAI-ROLL lends itself well to policy integration, as its human–AI bias mapping can be translated into auditable performance indicators. For example, it may leverage indicators such as a potential *Blind-Spot Score* to measure the share of out-of-distribution records detected during sampling tests, or a *Blind-Accept Rate* to track how often humans rubber-stamp automated outputs into post-deployment monitoring and audit protocols [4].

**Limitations and Potential uptake constraints,** should be recognized to institutionalizing HAI-ROLL in practice. Building fairness-aware AI systems requires more than technical soundness; it calls for practices that are sustainable and grounded in continuous reflection on the social and ethical impact of design choices. Cultural and organizational change, along with synchronized activities (educational programs, awareness campaigns, and cognitive bias-informed design protocols), would be needed to enable AI professionals and policy makers to agree on the roots of AI unfairness. Equally important is the promotion of inclusivity and diversity within AI development teams to counteract groupthink and bring forward perspectives that better detect and address bias [40]. `HAI-ROLL` offers a conceptualization framework and opens research dialogues to move away from current computationally-focused fairness solutions, by recognizing that AI models even carefully designed to be fair, might still remain biased due to human harms creeping at the overall AI lifecycle. The proposed framework is based on an extensive study of the most prominent state of the art evidence (from cognitive psychology, data science and AI) and rolls in the deeper understanding of the causes and effects of this human-AI confluence. `HAI-ROLL` and the proposed work can't be exhaustive.Rather, they offer a forward-looking and scientifically grounded foundation. The practical readiness of such bias-aware AI systems should be further and thoroughly validated through empirical studies. Its adoption would require domain-specific case studies scenarios and inter-disciplinary research consensus to practically verify its scientific evidence and advance its systematic mapping in real-world settings. As dicussed above, we claim that we offer a stepping stone to widen the study of human biases in FairAI by surfacing overlooked connections between cognitive heuristics and computational biases, and by highlighting meaningful patterns of influence and interdependencies. Our ongoing work on FairAI focuses on extending `HAI-ROLL` mapping with a holistic human-inclusive taxonomy which will be theoretically grounded and experimentally tested, under varying and bias-sensitive constraints posed in several critical FairAI domains. Future uptake steps should include empirical validation of each mapping proposition by designing experiments that compare decision-making under the influence of a cognitive heuristic to a condition in which no such biasing influence is present (i.e., a neutral baseline), similar to control-versus-treatment setups [20]. As an indicative starting point, we propose eliciting the heuristic through realistic design vignettes, logging the resulting design choices, and auditing the downstream model for the predicted computational bias—thereby providing converging evidence for "Heuristic → Harmful Action → AI Bias" chains in real-world scenarios and applications. By identifying the potential sources of bias at an early stage of the AI lifecycle, `HAI-ROLL` highly inspires the design of effective bias mitigation approaches. Our proposition does not prove that the hypotheses or assumptions of other researchers are wrong. In contrast, we leverage existing cognitive biases and FairAI scientific outcomes to bridge gaps and initiate a deeper exploration of the complex human-to-AI bias "iceberg".

## About the Authors

**Nicoleta Tantalaki** (tantalaki@csd.auth.gr) is a post-doctoral researcher at the Laboratory on Data and Web Science (Datalab), School of Informatics, Aristotle University of Thessaloniki, Greece.
**Athena Vakali** (avakali@csd.auth.gr) is a Professor at the School of Informatics, Aristotle University of Thessaloniki, Greece, where she leads the research Laboratory on Data and Web science (Datalab).